\documentclass[final]{cvpr}

\usepackage{times}
\usepackage{epsfig}
\usepackage{graphicx}
\usepackage{amsmath}
\usepackage{amssymb}
\usepackage[table, svgnames]{xcolor}

\usepackage[pagebackref=true,breaklinks=true,colorlinks,bookmarks=false]{hyperref}

\def\cvprPaperID{5317} 
\def\confYear{CVPR 2021}

\begin{document}

\title{A 3D GAN for Improved Large-pose Facial Recognition}


\author{Richard T. Marriott$^{1,2}$\qquad Sami Romdhani$^2$\qquad Liming Chen$^1$\\
{\normalsize
$^1$ Ecole Centrale de Lyon, France\qquad $^2$ IDEMIA, France}\\
{\tt\small richard.marriott@idemia.com}
}

\maketitle

\begin{abstract}
Facial recognition using deep convolutional neural networks relies on the availability of large datasets of face images. Many examples of identities are needed, and for each identity, a large variety of images are needed in order for the network to learn robustness to intra-class variation. In practice, such datasets are difficult to obtain, particularly those containing adequate variation of pose. Generative Adversarial Networks (GANs) provide a potential solution to this problem due to their ability to generate realistic, synthetic images. However, recent studies have shown that current methods of disentangling pose from identity are inadequate. In this work we incorporate a 3D morphable model into the generator of a GAN in order to learn a nonlinear texture model from in-the-wild images. This allows generation of new, synthetic identities, and manipulation of pose, illumination and expression without compromising the identity. Our synthesised data is used to augment training of facial recognition networks with performance evaluated on the challenging CFP and CPLFW datasets.
\end{abstract}

\section{Introduction}
\label{sec:intro}

State-of-the-art facial recognition (FR) algorithms are trained using millions of images. With the internet as a resource, face-images are relatively easy to come by. However, the distribution of semantics throughout these images is usually highly unbalanced. For example, the majority of available photographs are frontal portraits of smiling subjects, with images containing large poses being relatively scarce. Robustness to pose is currently considered to be one of the largest challenges for FR. Some researchers have attempted to avoid the problem by first frontalising probe images \cite{DBLP:conf/cvpr/HassnerHPE15, DBLP:conf/cvpr/Shen0YWT18, DBLP:conf/cvpr/ZhuLYYL15}, whilst others have attempted to learn additional robustness to pose by synthetically augmenting training datasets \cite{DBLP:journals/corr/CrispellBCBM17, DBLP:conf/cvpr/DengCXZZ18, DBLP:conf/eccv/MasiTHLM16, DBLP:conf/nips/ZhaoXKLZWPSYF17}. We advocate this second approach since it does not require additional resources during inference.

\begin{figure}
\begin{center}
\setlength\tabcolsep{0pt}
\renewcommand{\arraystretch}{0}
\begin{tabular}{ccccccc}
\includegraphics[width=0.142\linewidth]{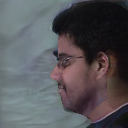}&\includegraphics[width=0.142\linewidth]{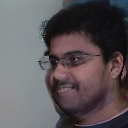}&\includegraphics[width=0.142\linewidth]{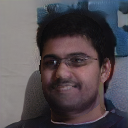}&\includegraphics[width=0.142\linewidth]{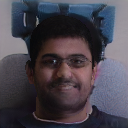}&\includegraphics[width=0.142\linewidth]{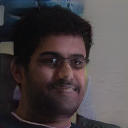}&\includegraphics[width=0.142\linewidth]{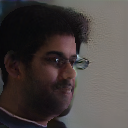}&\includegraphics[width=0.142\linewidth]{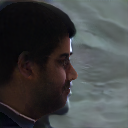}\\
\includegraphics[width=0.142\linewidth]{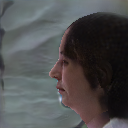}&\includegraphics[width=0.142\linewidth]{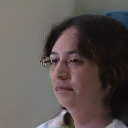}&\includegraphics[width=0.142\linewidth]{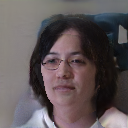}&\includegraphics[width=0.142\linewidth]{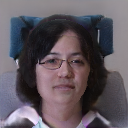}&\includegraphics[width=0.142\linewidth]{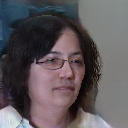}&\includegraphics[width=0.142\linewidth]{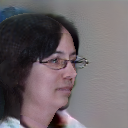}&\includegraphics[width=0.142\linewidth]{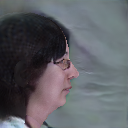}\\
\includegraphics[width=0.142\linewidth]{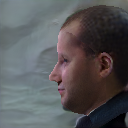}&\includegraphics[width=0.142\linewidth]{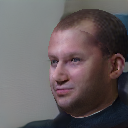}&\includegraphics[width=0.142\linewidth]{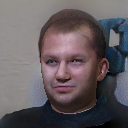}&\includegraphics[width=0.142\linewidth]{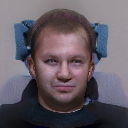}&\includegraphics[width=0.142\linewidth]{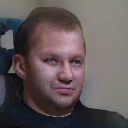}&\includegraphics[width=0.142\linewidth]{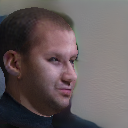}&\includegraphics[width=0.142\linewidth]{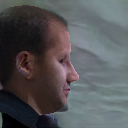}\\
\includegraphics[width=0.142\linewidth]{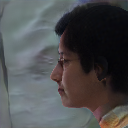}&\includegraphics[width=0.142\linewidth]{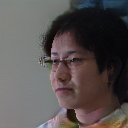}&\includegraphics[width=0.142\linewidth]{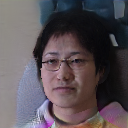}&\includegraphics[width=0.142\linewidth]{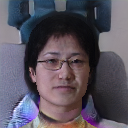}&\includegraphics[width=0.142\linewidth]{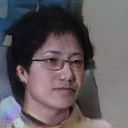}&\includegraphics[width=0.142\linewidth]{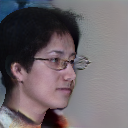}&\includegraphics[width=0.142\linewidth]{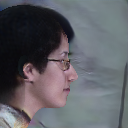}\\
\end{tabular}
\end{center}
   \caption{Instances of the FLAME 3D morphable model rendered at various poses with nonlinear textures (and background) learned by our 3D GAN from the MultiPIE dataset. (The model instances correspond to the \textit{Neutral} expression column of Figure \ref{fig:MultiPIE_Expression}.)}
\label{fig:MultiPIE_Pose}
\end{figure}

Synthetic augmentation of poses in training data has typically been achieved by fitting some 3D face model to input images, extracting textures, and then re-projecting those textures at modified poses \cite{DBLP:journals/corr/CrispellBCBM17, DBLP:conf/nips/ZhaoXKLZWPSYF17}. With recent advances in the development of Generative Adversarial Networks (GANs), however, a viable alternative has emerged. GANs have been shown to be capable of generating realistic images of new identities and so restricting data-augmentation to existing identities is not necessary. In order to generate fully synthetic training data, however, disentanglement of identity from other characteristics, such as pose, is necessary. Recent studies have shown that 2D GAN methods struggle to achieve this disentanglement \cite{marriott2020assessment}. In this work we incorporate a 3D morphable model (3DMM) \cite{DBLP:journals/tog/LiBBL017} into a GAN so that images of new, synthetic identities can be generated, and the pose modified without identity being compromised. Our contributions are:

\begin{enumerate}
    \item Introduction of a method of learning a nonlinear texture model from \textit{in-the-wild images} that can be used to generate images of \textit{synthetic identities} with fully disentangled pose. No specially captured scans of facial texture are required.
    \item Demonstration of improvements to large-pose facial recognition by augmenting datasets with synthetic, 3D GAN images, and a state-of-art accuracy for CPLFW.
\end{enumerate}


\section{Related Work}
\label{sec:related work}

\subsection{Generative 3D networks}
\label{sec: RW Generative 3D networks}


Prior to the recent explosion in the development of GAN-related methods, the best way of generating synthetic face images was to use a 3D morphable model (3DMM) \cite{DBLP:conf/siggraph/BlanzV99, DBLP:conf/cvpr/BoothRZPD16, DBLP:conf/cvpr/TranL019}. These models are generally built from a limited number of 3D shape and texture scans. To achieve greater representativeness, methods such as \cite{DBLP:conf/cvpr/BoothAPTPZ17} were proposed in which PCA-based texture models are learnt from large datasets of in-the-wild images. While the linear spaces of these models are known to capture most of the variation in the training datasets, generated faces still appear to be smooth with textures lacking in high frequency detail. In \cite{DBLP:conf/eccv/GecerLPDPMZ20} and \cite{DBLP:conf/cvpr/GecerPKZ19} the linear texture model of the LSFM \cite{DBLP:conf/cvpr/BoothRZPD16} is replaced by the nonlinear, CNN generator of a GAN trained to approximate the distribution of their dataset of high-quality texture scans. The quality of generated textures is outstanding. However, the dataset of scans is not available for general use.

The difficulty of obtaining high-quality texture datasets motivates the development of methods such as our own, which aims to learn a nonlinear texture model from natural (non-scanned) images. Various other methods have a similar aim and attempt to train disentangled auto-encoders to learn texture and shape models via reconstruction of in-the-wild training images \cite{DBLP:conf/eccv/ChaudhuriVSW20, DBLP:conf/cvpr/TewariB0BESPZT19, DBLP:conf/cvpr/TewariZ0BKPT18, DBLP:conf/cvpr/TranL019, DBLP:conf/cvpr/Tran018}. New identities, however, can only be generated by empirically exploring the latent space of their generators. The method we propose is a GAN rather than an auto-encoder, and so can more reliably and straight-forwardly generate new, synthetic identities.

A number of recent methods have attempted to build upon the success of GANs in the generation of high-quality images by integrating 3D models to gain control over semantics. In \cite{DBLP:conf/cvpr/TewariEBBSPZT20} and \cite{DBLP:journals/tog/TewariE0BSPZT20}, mappings from 3DMM parameters to the latent space of a pre-trained StyleGAN2 \cite{DBLP:conf/cvpr/KarrasLAHLA20} are learned such that the similarity of generated images with 3DMM renderings is maximised. In \cite{DBLP:conf/3dim/GhoshGURBB20}, \cite{DBLP:conf/eccv/KowalskiGEBJS20} and \cite{DBLP:conf/eccv/GecerBKK18}, GANs are trained to generate realistic images but are conditioning in various ways on renderings of 3D models. For each of these methods, generation takes place in image-space rather than texture-space meaning that appearance and identity are not implicitly robust to pose transformations. Pre-trained biometrics networks are therefore often applied as losses to constrain identity. The quality of these biometric networks imposes an upper limit on the usefulness of synthesised data for the application of data-augmentation for FR.


\subsection{Large-pose facial recognition}
\label{sec: RW Large-pose data-aug}


Large-pose FR could be framed as a problem of limited data or as a problem of how to better learn from available data. Various generic methods have been shown to perform well on large-pose FR benchmarks \cite{DBLP:conf/cvpr/HuangWT0SLLH20, DBLP:conf/eccv/HuangSTL0LHJ20}. Other methods specifically disentangle pose from identity representations by regressing/classifying pose labels \cite{DBLP:conf/cvpr/CaoR0TL18,DBLP:conf/iccv/PengYSMC17,DBLP:conf/cvpr/Tran0017}. Complementary to these works are methods for synthetically augmenting pose in training images. Earlier methods involved extraction of textures from images onto the surface of a 3D model for manipulation of pose, and sometimes illumination or expression \cite{DBLP:journals/corr/CrispellBCBM17, DBLP:journals/ijon/LvSHZZ17, DBLP:conf/eccv/MasiTHLM16, DBLP:conf/iccv/PengYSMC17}. Due to self-occlusion in images and therefore holes in the textures, various in-filling techniques were employed. In \cite{DBLP:conf/nips/ZhaoXKLZWPSYF17} this problem is tackled by refining the projected texture in image-space using an adversarial loss. This final refinement phase requires an identity-preserving loss that, as previously mentioned, limits the usefulness of the synthetic data for improving FR.

A preferable method is to produce a complete texture in the texture reference space of the 3D model, thus ensuring that the identity remains consistent when projected to different poses. In \cite{DBLP:conf/cvpr/DengCXZZ18}, a texture-completion network is trained using a set of carefully prepared ground-truth textures. In \cite{DBLP:conf/eccv/GecerLPDPMZ20} and \cite{DBLP:journals/corr/abs-1802-05891} the problem of texture completion is avoided entirely by generating textures for \textit{synthetic} identities.  \cite{DBLP:journals/corr/abs-1802-05891} uses a linear texture model whereas \cite{DBLP:conf/eccv/GecerLPDPMZ20} trains a nonlinear model. Each of these methods makes use of datasets of scanned textures. The method proposed here also generates synthetic identities in order to avoid the problems of texture completion and reconstruction of existing identities. The method, however, does not require carefully prepared/captured ground-truth textures and, instead, learns a nonlinear texture model directly from in-the-wild images.

\begin{figure*}
\begin{center}
\includegraphics[width=0.9\linewidth]{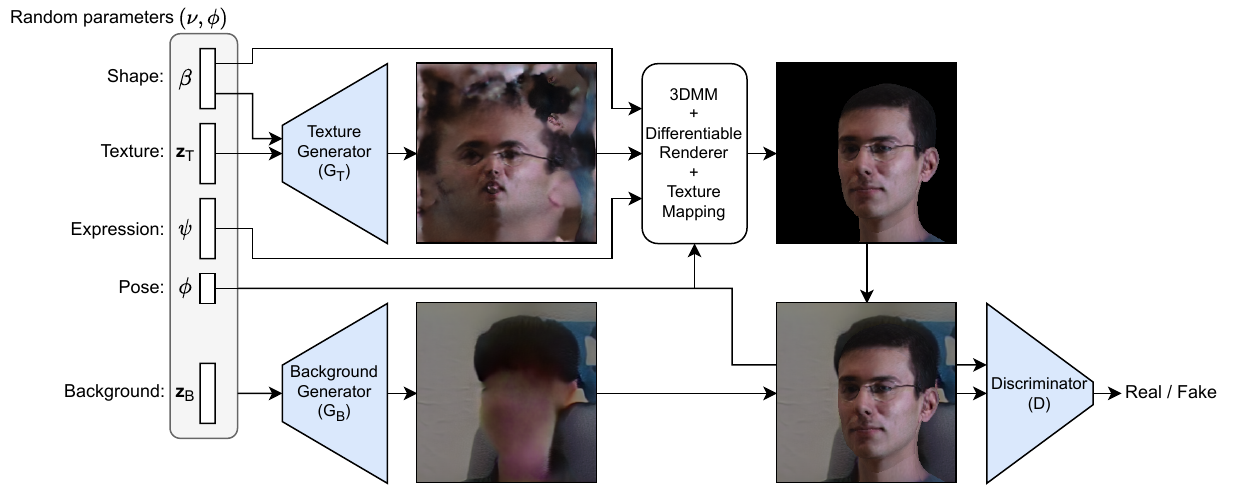}
\end{center}
   \caption{The 3D GAN's generator includes two CNNs that generate facial texture and background. Facial texture is rendered into the background using some random sample of shape from the 3D model's distribution. The random pose and expression vectors are used only for rendering, not for generation of texture, and so remain disentangled from the identity. All parameters are passed to the background generator to allow harmonisation of the background conditions with the rendered subject. Note that all vectors are randomly sampled and that no direct comparison with training images is performed.}
\label{fig:3DGAN}
\end{figure*}

\section{The 3D GAN}
\label{sec:3DGAN}

Generative Adversarial Networks (GANs) typically consist of a convolutional generator and discriminator that are trained alternately in a mini-max game: the discriminator is trained to distinguish generated images from those of a training set of real images, and the generator is trained to minimise the success of the discriminator. Although generated images appear to represent real-world, 3D subjects, they are of course nothing more than combinations of 2D features learned by the 2D convolutional filters of the generator. For this reason, upon linearly traversing the latent space of a GAN's generator, one tends to see ``lazy'', 2D transformations between forms rather than transformations that would be physically realistic in a 3D space. For example, even if a direction in the latent space is identified that influences the pose of a face in a generated image, the appearance of the face is unlikely to remain consistent. Indeed, the generator may not even be capable of generating the same face at a different pose. In order to ensure that appearance (and identity) is maintained in synthesised images upon manipulation of pose, we enhance the generator by integrating a 3DMM.

\pagebreak
Typically, input to a GAN is a random vector. The inputs to our 3D GAN are random texture and background vectors, but also random 3DMM shape, expression and pose parameters. A differentiable renderer is then used to render random head-shapes into a generated ``background image'' with the facial texture being provided by a texture generator that learns its model from scratch from the training images. No matter what the shape, expression or pose of the random model instance, the rendered image must appear realistic to the discriminator. To achieve this, the texture generator learns to generate realistic textures with features that correctly correspond with the model shape.

Figure \ref{fig:3DGAN} depicts the architecture of our 3D GAN. The lower half of the diagram depicts a standard conditional GAN in which some image is generated from random parameters, and is then fed to the discriminator along with the pose parameters to correspond with training images also labelled with poses. In our implementation, pose information is repeated spatially and concatenated as additional channels of the image. The top half of the diagram depicts the integration of a 3DMM where a learned texture is rendered into this image via a differentiable renderer. With the main subject of the image being provided by the rendered texture, the background generator learns to generate only the background and features not modelled by the 3DMM, for example, the edges of glasses, clothes and hair. Since the texture generator is not conditioned on pose, nor on expression parameters, these aspects of the image can be manipulated without affecting the texture of the 3D model, as shown in Figures \ref{fig:MultiPIE_Pose}, \ref{fig:MultiPIE_Expression} and \ref{fig:FFHQ_random}. There is therefore no need for an additional identity constraint.

\subsection{Implementation}
\label{sec:Implementation}

\begin{figure}
\begin{center}
\begin{tabular}{ccc}
\includegraphics[width=0.25\linewidth]{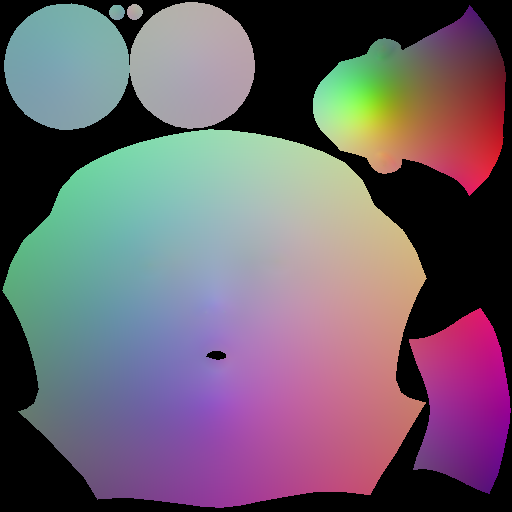} & \includegraphics[width=0.25\linewidth]{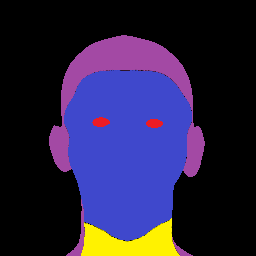} & \includegraphics[width=0.25\linewidth]{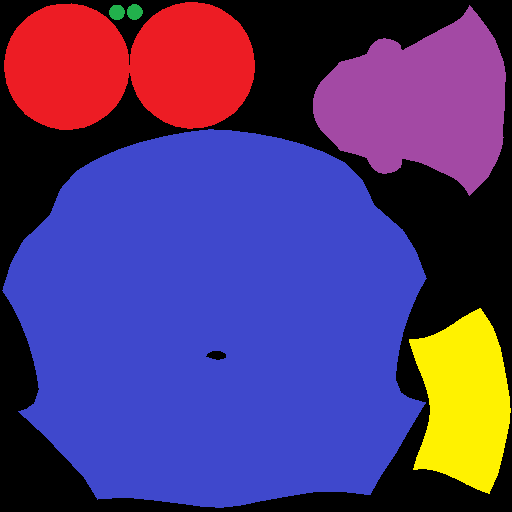}\\
(a) & (b) & (c)
\end{tabular}
\end{center}
   \caption{a) The FLAME 3DMM's texture map where RGB represents the corresponding 3D point on the mean model shape; b) a rendering of the texture shown in (c).}
\label{fig:Texture_map}
\end{figure}

Our full generator is a function of five sets of random input parameters and two sets of trained parameters:
\begin{align}
    \mathbf{x} = &G([\mathbf{z}_T, \mathbf{z}_B, \mathbf{\beta}, \mathbf{\psi}, \mathbf{\phi}]; [\mathbf{\theta}_T, \mathbf{\theta}_B]) \label{eq:G_params} \\
    = &(\mathbf{1} - \mathbf{K}) \circ G_B(\mathbf{z}_B, \mathbf{z}_T, \mathbf{\beta}, \mathbf{\psi}, \mathbf{\phi}; \mathbf{\theta}_B) \nonumber \\
    &+ \mathbf{K} \circ \mathcal{M}(G_T(\mathbf{z}_T, \mathbf{\beta}; \mathbf{\theta}_T), \mathbf{y}) \label{eq:G_expanded}
\end{align}
where $\mathbf{x}$ is a generated image; $G_B$ and $G_T$ are the background and texture generators; $\mathbf{z}_T \in \mathcal{N}^{N_T}$ and $\mathbf{z}_B \in \mathcal{N}^{N_B}$ are vectors of random texture and background parameters of length $N_T$ and $N_B$ respectively, selected from standard normal distributions; $\mathbf{\beta} \in \mathcal{N}^{N_s}$ and $\mathbf{\psi} \in \mathcal{N}^{N_e}$ are vectors of shape and expression parameters that control the form of the 3DMM, again selected from standard normal distributions; $\mathbf{\phi}$ is pose information, typically values of yaw and pitch selected at random from the labels of the training set of images; and $\mathbf{\theta}_T$ and $\mathbf{\theta}_B$ parametrise the texture and background generator networks. The background image and rendered texture are combined using a binary mask, $\mathbf{K}$, generated by the renderer. Note that the masking by $\mathbf{K}$ is not shown in Figure \ref{fig:3DGAN}. $\mathbf{1}$ is a vector of ones of the same shape as the image and $\mathbf{a} \circ \mathbf{b}$ represents the element-wise product of vectors $\mathbf{a}$ and $\mathbf{b}$. $\mathcal{M}$ is an inverse texture-mapping function that maps interpolations from the generated texture map to appropriate locations in image space based on a rendering of texture coordinates in image-space, $\mathbf{y}$. Inverse texture mapping effectively allows the generated texture to be pasted onto the model surface rather than having only single colours at each vertex and interpolation across facets. Our texture generator operates at twice the resolution of the background generator to help avoid pixelation of surfaces that are less well represented in the texture map. Rendering of $\mathbf{y}$ (and simultaneously, $\mathbf{K}$) is performed by the differentiable rendering function, $R$:
\begin{align}
    \mathbf{y}, \mathbf{K} = R(\mathbf{S}, \mathbf{\phi}; \mathbf{\tau}, \mathbf{\gamma}) \label{eq:Renderer}
\end{align}
where $\mathbf{S} \in \mathbb{R}^{N_v \times 3}$ is a vector of shape vertices for some random instance of the 3DMM; $\phi$ is pose information; $\mathbf{\tau} \in \mathbb{Z}^{N_{\tau} \times 3}$ is the 3DMM's triangle list of $N_{\tau}$ vertex indices; and $\mathbf{\gamma} \in \mathbb{R}^{3N_{\tau} \times 2}$ is the vector of texture coordinates where each of the $N_{\tau}$ triangles has its own set of three 2D texture vertices. The rendering function, $R$, is implemented by DIRT (Differentiable Renderer for Tensorflow) \cite{DBLP:journals/ijcv/HendersonF20} and we use the FLAME (Faces Learned with an Articulated Model and Expressions) \cite{DBLP:journals/tog/LiBBL017} 3DMM. FLAME is an articulated model with joints controlling the head position relative to the neck, the gaze direction, and the jaw. To avoid having to estimate of the distributions of these angles, during training of our 3D GAN we fix the joint parameters such that the shape is given by the following, simplified equation
\begin{align}
    \mathbf{S} = \bar{\mathbf{S}} + \sum_{n=1}^{N_s} b_n \mathbf{s}_n +  \sum_{n=1}^{N_e} c_n \mathbf{e}_n \label{eq:3DMM}
\end{align}
where $\bar{\mathbf{S}}$ is the mean model shape; $\mathbf{\mathcal{S}} = [\mathbf{s}_1, ..., \mathbf{s}_{N_s}]$ are the principal components of shape; $\mathbf{\epsilon} = [\mathbf{e}_1, ..., \mathbf{e}_{N_e}]$ are the principal components of expression; and $[b_1, ..., b_{N_s}]$ and $[c_1, ..., c_{N_e}]$ are the individual elements of the previously defined shape and expression vectors, $\mathbf{\beta}$ and $\mathbf{\psi}$, that are also fed to the generator networks in Equation (\ref{eq:G_expanded}). For the FLAME model, $N_s = 200$, $N_e = 200$, $N_v = 5023$ and $N_{\tau} = 9976$. We also set $N_T = N_B = 200$.

The architectures of $G_T$ and $G_B$ are based on that of the Progressive GAN \cite{DBLP:conf/iclr/KarrasALL18}. However, to simplify implementation and speed up training, no progressive growing was used. We believe that use of a 3D model may act to stabilise training since it provides prior form that need not be learned from scratch. The architecture was augmented with bilinear interpolation on upscaling (rather than nearest-neighbour upscaling), which helps to avoid checker-board artefacts, and with static Gaussian noise added to each feature map, as used in \cite{DBLP:conf/cvpr/KarrasLA19}, which helps to prevent wave-like artefacts from forming. (See Figure \ref{fig:Ablation_study} for examples.)

\subsection{Training}

Despite the more elaborate architecture of the generator, the 3D GAN can be trained like any other GAN. We choose to optimise a Wasserstein loss \cite{DBLP:conf/icml/ArjovskyCB17} by alternately minimising the discriminator and generator losses in Equations (\ref{eq:D_loss}) and (\ref{eq:G_loss}) respectively. The values of all input vectors (with the exception of the conditional pose parameters) are selected from a standard Gaussian distribution. For simplicity of notation, we agglomerate them into a single vector $\mathbf{\nu} = [\mathbf{z}_T, \mathbf{z}_B, \mathbf{\beta}, \mathbf{\psi}]$.
\begin{align}
    \mathcal{L}_{\theta_D} = &\mathbb{E}_{(\mathbf{x}_r, \mathbf{\phi})\sim p_{data}}[D(\mathbf{x}_r, \mathbf{\phi}; \theta_D)] \nonumber \\
    &- \mathbb{E}_{\mathbf{\nu}\sim\mathcal{N}, (\_, \mathbf{\phi})\sim p_{data}}[D(G(\mathbf{\nu}, \mathbf{\phi}; \theta_G), \mathbf{\phi}; \theta_D)] + Reg. \label{eq:D_loss} \\
    \mathcal{L}_{\theta_G} = &\mathbb{E}_{\mathbf{\nu}\sim\mathcal{N}, (\_\mathbf{\phi})\sim p_{data}}[D(G(\mathbf{\nu}, \mathbf{\phi}; \theta_G), \mathbf{\phi}; \theta_D)] \label{eq:G_loss}
\end{align}
where $D$ is the discriminator network parametrised by $\theta_D$; $(\mathbf{x}_r, \phi)$ is a real image and associated pose labels selected at random from the distribution of training data, $p_{data}$; $\theta_G = [\theta_T, \theta_B]$; and $Reg.$ indicates the addition of a gradient penalty \cite{DBLP:conf/nips/GulrajaniAADC17} that acts to regularise the discriminator such that it approximately obeys the required k-Lipschitz condition \cite{DBLP:conf/icml/ArjovskyCB17}. Note that, during training, the shape and expression parameters passed to the generator are random. There is never any direct reconstruction of training images via fitting of the 3D model. The only constraint on textures is that they must appear realistic (as judged by the discriminator) when projected at any angle and with any expression. The motivation for training our generator as a GAN and avoiding reconstruction losses is to be able to generate new identities and to avoid generation of smoothed textures caused by reconstruction errors.

\subsection{Limitations}
\label{sec:Limitations}

The 3D GAN method has certain limitations, the most fundamental possibly being the fact that hair, glasses and teeth are not included in the 3D shape model. This can lead to projections of these features onto the surface of the model that do not necessarily look realistic when viewed from certain angles. The inclusion of such features in the shape model would be difficult at best. Instead, it may be better to detect and remove images containing unmodelled features from the training dataset and to seek another method for augmentation with glasses and occlusion by overhanging hair, for example \cite{DBLP:journals/ijon/LvSHZZ17}.

As currently formulated, the 3D GAN learns lighting effects and shadows as part of the texture. Although this helps generated images appear to be realistic, it is not ideal for our goal of improving FR since specific lighting conditions become part of the synthetic identities. Since we have the 3D shape for each generated image, a lighting model could be used to produce shading maps of randomised lighting conditions during training similar to that performed in \cite{DBLP:conf/cvpr/Tran018}. Ideally, the random lighting conditions should follow the distribution of lighting in the training set. In this way the texture generator might avoid inclusion of the modelled lighting effects in the texture.

We also make the assumption that the distributions of shape and expression in the training dataset match the natural distributions of the 3DMM. This is not necessarily the case and improvements could be possible by first fitting the model to the dataset. N.B. we suggest this only for estimating the distributions, not for reconstructing images since fitting errors would be large in individual cases. We also assume that the distributions of feature points (used for alignment) and poses are known.

\begin{figure}
\begin{center}
\setlength\tabcolsep{0pt}
\renewcommand{\arraystretch}{0}
\begin{tabular}{ccccccc}
\includegraphics[width=0.142\linewidth]{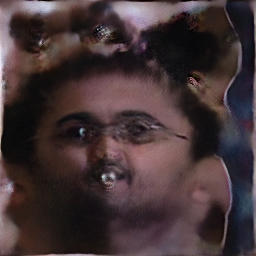}&\includegraphics[width=0.142\linewidth]{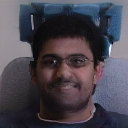}&\includegraphics[width=0.142\linewidth]{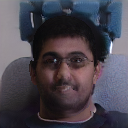}&\includegraphics[width=0.142\linewidth]{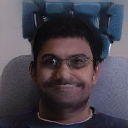}&\includegraphics[width=0.142\linewidth]{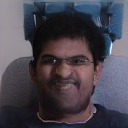}&\includegraphics[width=0.142\linewidth]{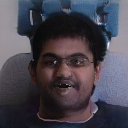}&\includegraphics[width=0.142\linewidth]{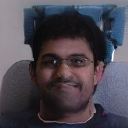}\\
\includegraphics[width=0.142\linewidth]{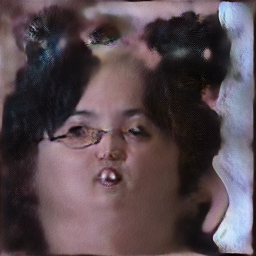}&\includegraphics[width=0.142\linewidth]{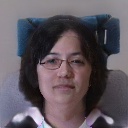}&\includegraphics[width=0.142\linewidth]{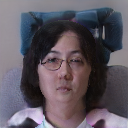}&\includegraphics[width=0.142\linewidth]{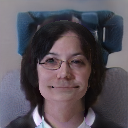}&\includegraphics[width=0.142\linewidth]{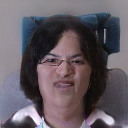}&\includegraphics[width=0.142\linewidth]{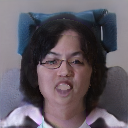}&\includegraphics[width=0.142\linewidth]{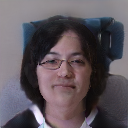}\\
\includegraphics[width=0.142\linewidth]{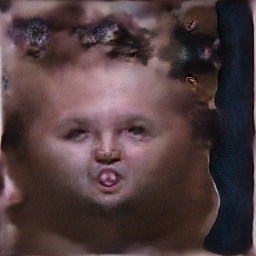}&\includegraphics[width=0.142\linewidth]{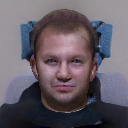}&\includegraphics[width=0.142\linewidth]{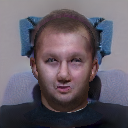}&\includegraphics[width=0.142\linewidth]{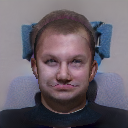}&\includegraphics[width=0.142\linewidth]{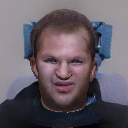}&\includegraphics[width=0.142\linewidth]{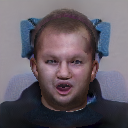}&\includegraphics[width=0.142\linewidth]{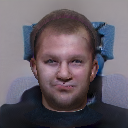}\\
\includegraphics[width=0.142\linewidth]{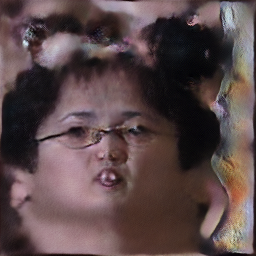}&\includegraphics[width=0.142\linewidth]{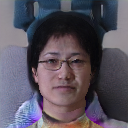}&\includegraphics[width=0.142\linewidth]{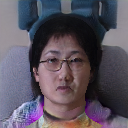}&\includegraphics[width=0.142\linewidth]{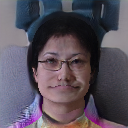}&\includegraphics[width=0.142\linewidth]{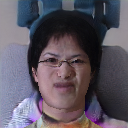}&\includegraphics[width=0.142\linewidth]{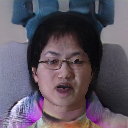}&\includegraphics[width=0.142\linewidth]{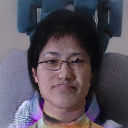}\\[0.1cm]
Texture & Neutral & Exp 1 & Exp 2 & Exp 3 & Exp 4 & Exp 5
\end{tabular}
\end{center}
   \caption{Renderings of the FLAME morphable model for various expressions with textures learned by the 3D GAN from the Multi-PIE dataset. Output of the texture generator prior to rendering is shown in the first column.}
\label{fig:MultiPIE_Expression}
\end{figure}

Finally, the texture map provided with the FLAME 3DMM (see Figure \ref{fig:Texture_map}a) is spatially discontinuous. Since CNNs function by exploiting spatial coherence, these discontinuities in the texture-space lead to discontinuity artefacts in the rendered images. This can be seen, for example, in Figure \ref{fig:MultiPIE_Expression} where the facial texture meets the texture of the back of the head. These artefacts could largely be avoided by using an alternative, spatially continuous texture mapping.

\section{Results}
\label{sec:Results}

\subsection{Qualitative evaluation for a controlled dataset}
\label{sec:Results_Development}

During development of the 3D GAN, tests were conducted by training on the controlled, Multi-PIE dataset \cite{DBLP:journals/ivc/GrossMCKB10}. Doing so avoided potential problems that might have arisen due to incorrect estimation of poses, which are required to condition the GAN. During these tests, the pitch of the model was not varied and so we excluded Multi-PIE's CCTV-like camera angles (8 and 19). The first column of Figure \ref{fig:MultiPIE_Expression} shows examples of random textures learned by the 3D GAN. To demonstrate the level of correspondence with the shape model, we render each texture for six different expressions. We see that features are well aligned and that expressions can be manipulated realistically. This is thanks to the requirement that the texture look realistic for renderings of all poses and expressions. The texture is not dependent on the expression parameters and so the identity is implicitly maintained, at least to the limit of disentanglement present in the 3DMM. Figure \ref{fig:MultiPIE_Pose} shows renderings of the same textures with a neutral expression at a selection of yaw angles in the range $[-90^{\circ}, 90^{\circ}]$. We see that the model heads are pleasingly integrated with the background with additional, unmodelled details such as hair and the edges of glasses being generated. In some cases, however, this is problematic. For example, in the final column, something resembling a protruding chin has been generated in the background for both of the male subjects. Note, however, that the background is only needed for training and that facial textures can be rendered onto arbitrary backgrounds.

\begin{figure}
\begin{center}
\begin{tabular}{ccc}
\includegraphics[width=0.175\linewidth]{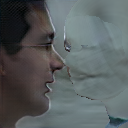} & \includegraphics[width=0.175\linewidth]{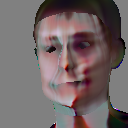} &
\includegraphics[width=0.175\linewidth]{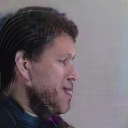} \\
a) No conditioning & b) No cond / bckd & c) No bilinear
\end{tabular}
\begin{tabular}{ccc}
\includegraphics[width=0.175\linewidth]{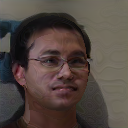} &
\includegraphics[width=0.175\linewidth]{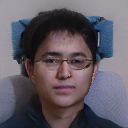} & \includegraphics[width=0.175\linewidth]{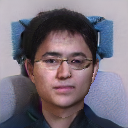}\includegraphics[width=0.175\linewidth]{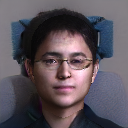} \\
d) No noise & e) Final & f) Final + SH lighting
\end{tabular}
\end{center}
   \caption{Results characterising the effects of various features of the final implementation of our 3D GAN.}
\label{fig:Ablation_study}
\end{figure}

\begin{figure*}[t]
    \centering
\includegraphics[width=0.42\linewidth]{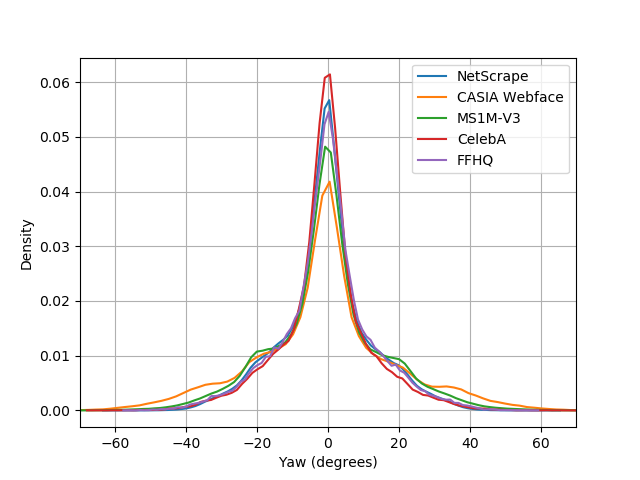}\includegraphics[width=0.42\linewidth]{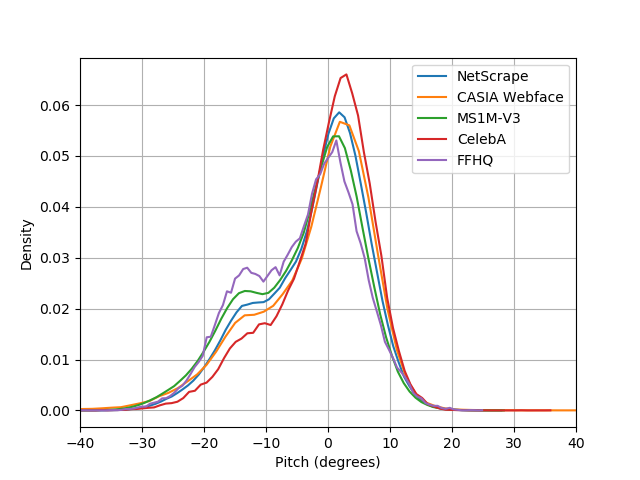}
   \caption{The relative pose distributions estimated for the non-synthetic datasets described in Section \ref{sec:Training sets} and Table \ref{tab:Dataset_comparison}.}
    \label{fig:Pose_dists}
\end{figure*}

Figure \ref{fig:Ablation_study} shows a set of images that characterise the effects of disabling/enabling various aspects of our 3D GAN. Figure \ref{fig:Ablation_study}a shows that disabling the pose-conditioning can lead to degenerate solutions where the generators conspire to generate faces as part of the background and to camouflage the model. In the given example, pose-conditioning would have caused the discriminator to expect a subject facing towards the left and to therefore penalise such an image. Attempting to avoid this problem by switching off the background generator causes a different problem that can be seen in Figure \ref{fig:Ablation_study}b. The texture generator now produces a mixture of face-like and background-like features in order to satisfy the discriminator. Figure \ref{fig:Ablation_study}c has the background and pose-conditioning enabled. It demonstrates, however, obvious checker-board artefacts in the texture. We found that this problem was caused by the nearest-neighbour up-sampling of feature-maps upon resolution doubling within the generator. Following the work of \cite{DBLP:conf/cvpr/KarrasLA19}, we switched to bilinear up-sampling. Whilst this prevented the checkerboard artefacts, it led to the wave-like artefacts that can be seen in Figure \ref{fig:Ablation_study}d. Finally, in Figures \ref{fig:Ablation_study}e and \ref{fig:Ablation_study}f, we added static, Gaussian noise into the generator \cite{DBLP:conf/cvpr/KarrasLA19}. The noise acts to provide high-frequency, stochastic features by default so that the generator need not attempt to derive these details from the random input vectors. In Figure \ref{fig:Ablation_study}f we have also enabled spherical harmonic (SH) lighting \cite{DBLP:conf/siggraph/RamamoorthiH01a}, which is used in our data-augmentation experiments.

\begin{table}[t]
\begin{center}
\resizebox{\columnwidth}{!}{%
\begin{tabular}{|l|l|l|c|c|}
\hline
Dataset & IDs & Images & FR & GAN \\
\hline\hline
MS1M-V3 \cite{DBLP:journals/corr/abs-1905-00641} & 93.4k & 5.2M & \checkmark & \\
NetScrape (in-house) & 26.8k & 3.5M & \checkmark & \\
CASIA Webface \cite{DBLP:journals/corr/YiLLL14a} & 10.6k & 0.5M & \checkmark & \\
CelebA \cite{DBLP:conf/iccv/LiuLWT15} & 10.2k & 0.2M & \checkmark & \checkmark \\
Flickr-Faces-HQ \cite{DBLP:conf/cvpr/KarrasLA19} & N/A & 0.07M & & \checkmark \\
\hline
\end{tabular}
}
\end{center}
\caption{Comparison of datasets used in our data-augmentation experiments to train FR networks and/or the 3D GAN.}
\label{tab:Dataset_comparison}
\end{table}

\subsection{Data-augmentation in the wild}
\label{sec:Results_DataAug}

In the previous section we saw that it is possible to learn textures of good quality from a controlled dataset of images containing a wide range of pose. It is unlikely, however, that the synthetic 3D GAN data would be more informative than the original, high-quality dataset. Although the 3D GAN is able to generate new identities and allows full control over the pose, the data also inevitably suffers from problems such as mode-collapse and from limited realism. In this section we demonstrate improvement to FR by making better use of noisy, in-the-wild datasets. We present experiments for various FR algorithms trained on a variety of datasets. Evaluation was performed for two challenging, large-pose datasets: Celebrities in Frontal-Profile in the Wild (CFP) \cite{DBLP:conf/wacv/SenguptaCCPCJ16} and Cross-Pose LFW (CPLFW) \cite{zheng2018cross}, as well as their frontal-frontal counterparts. Benefit from use of 3D GAN data arises from a combination of the balanced distribution of poses and expressions, the use of a 3D lighting model, the presence of additional synthetic identities, and the GAN's ability to ``clean'' noisy datasets.

\begin{figure*}[t]
    \centering
    \renewcommand{\arraystretch}{0}
    \setlength{\tabcolsep}{12pt}
\begin{tabular}{cc}
\includegraphics[width=0.06\linewidth]{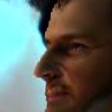}\includegraphics[width=0.06\linewidth]{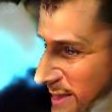}\includegraphics[width=0.06\linewidth]{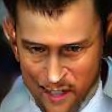}\includegraphics[width=0.06\linewidth]{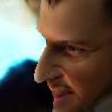}\includegraphics[width=0.06\linewidth]{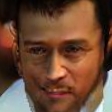}\includegraphics[width=0.06\linewidth]{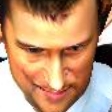}\includegraphics[width=0.06\linewidth]{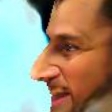}
&
\includegraphics[width=0.06\linewidth]{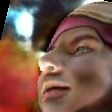}\includegraphics[width=0.06\linewidth]{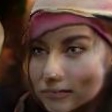}\includegraphics[width=0.06\linewidth]{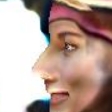}\includegraphics[width=0.06\linewidth]{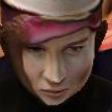}\includegraphics[width=0.06\linewidth]{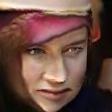}\includegraphics[width=0.06\linewidth]{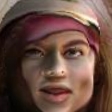}\includegraphics[width=0.06\linewidth]{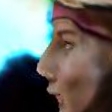} \\

\includegraphics[width=0.06\linewidth]{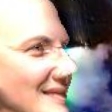}\includegraphics[width=0.06\linewidth]{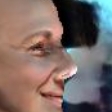}\includegraphics[width=0.06\linewidth]{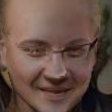}\includegraphics[width=0.06\linewidth]{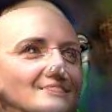}\includegraphics[width=0.06\linewidth]{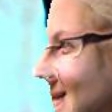}\includegraphics[width=0.06\linewidth]{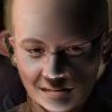}\includegraphics[width=0.06\linewidth]{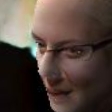}
&
\includegraphics[width=0.06\linewidth]{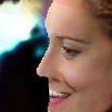}\includegraphics[width=0.06\linewidth]{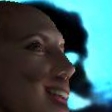}\includegraphics[width=0.06\linewidth]{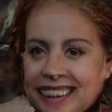}\includegraphics[width=0.06\linewidth]{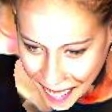}\includegraphics[width=0.06\linewidth]{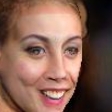}\includegraphics[width=0.06\linewidth]{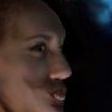}\includegraphics[width=0.06\linewidth]{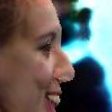} \\

\includegraphics[width=0.06\linewidth]{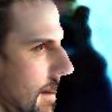}\includegraphics[width=0.06\linewidth]{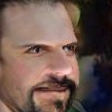}\includegraphics[width=0.06\linewidth]{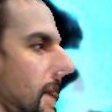}\includegraphics[width=0.06\linewidth]{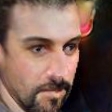}\includegraphics[width=0.06\linewidth]{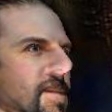}\includegraphics[width=0.06\linewidth]{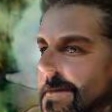}\includegraphics[width=0.06\linewidth]{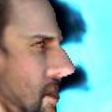}
&
\includegraphics[width=0.06\linewidth]{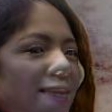}\includegraphics[width=0.06\linewidth]{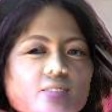}\includegraphics[width=0.06\linewidth]{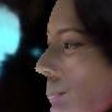}\includegraphics[width=0.06\linewidth]{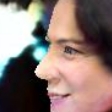}\includegraphics[width=0.06\linewidth]{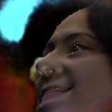}\includegraphics[width=0.06\linewidth]{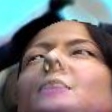}\includegraphics[width=0.06\linewidth]{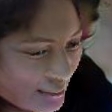} \\

\includegraphics[width=0.06\linewidth]{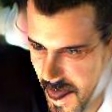}\includegraphics[width=0.06\linewidth]{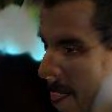}\includegraphics[width=0.06\linewidth]{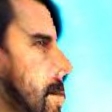}\includegraphics[width=0.06\linewidth]{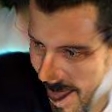}\includegraphics[width=0.06\linewidth]{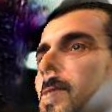}\includegraphics[width=0.06\linewidth]{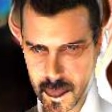}\includegraphics[width=0.06\linewidth]{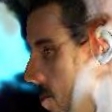}
&
\includegraphics[width=0.06\linewidth]{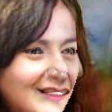}\includegraphics[width=0.06\linewidth]{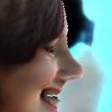}\includegraphics[width=0.06\linewidth]{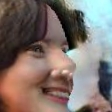}\includegraphics[width=0.06\linewidth]{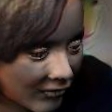}\includegraphics[width=0.06\linewidth]{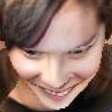}\includegraphics[width=0.06\linewidth]{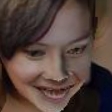}\includegraphics[width=0.06\linewidth]{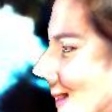} \\

\includegraphics[width=0.06\linewidth]{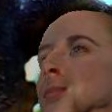}\includegraphics[width=0.06\linewidth]{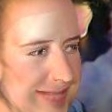}\includegraphics[width=0.06\linewidth]{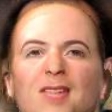}\includegraphics[width=0.06\linewidth]{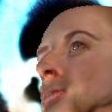}\includegraphics[width=0.06\linewidth]{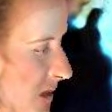}\includegraphics[width=0.06\linewidth]{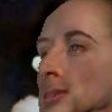}\includegraphics[width=0.06\linewidth]{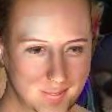}
&
\includegraphics[width=0.06\linewidth]{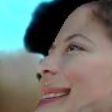}\includegraphics[width=0.06\linewidth]{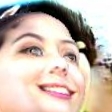}\includegraphics[width=0.06\linewidth]{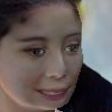}\includegraphics[width=0.06\linewidth]{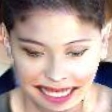}\includegraphics[width=0.06\linewidth]{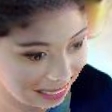}\includegraphics[width=0.06\linewidth]{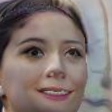}\includegraphics[width=0.06\linewidth]{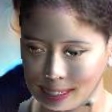} \\

\includegraphics[width=0.06\linewidth]{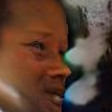}\includegraphics[width=0.06\linewidth]{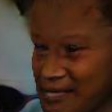}\includegraphics[width=0.06\linewidth]{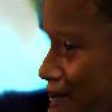}\includegraphics[width=0.06\linewidth]{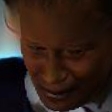}\includegraphics[width=0.06\linewidth]{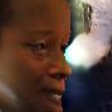}\includegraphics[width=0.06\linewidth]{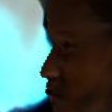}\includegraphics[width=0.06\linewidth]{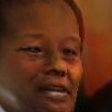}
&
\includegraphics[width=0.06\linewidth]{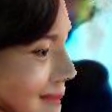}\includegraphics[width=0.06\linewidth]{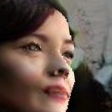}\includegraphics[width=0.06\linewidth]{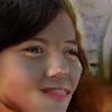}\includegraphics[width=0.06\linewidth]{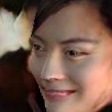}\includegraphics[width=0.06\linewidth]{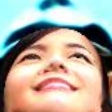}\includegraphics[width=0.06\linewidth]{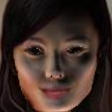}\includegraphics[width=0.06\linewidth]{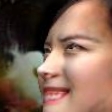} \\

\includegraphics[width=0.06\linewidth]{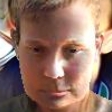}\includegraphics[width=0.06\linewidth]{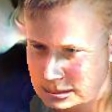}\includegraphics[width=0.06\linewidth]{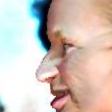}\includegraphics[width=0.06\linewidth]{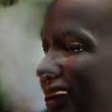}\includegraphics[width=0.06\linewidth]{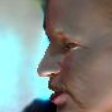}\includegraphics[width=0.06\linewidth]{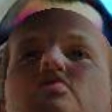}\includegraphics[width=0.06\linewidth]{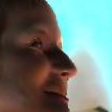}
&
\includegraphics[width=0.06\linewidth]{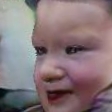}\includegraphics[width=0.06\linewidth]{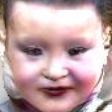}\includegraphics[width=0.06\linewidth]{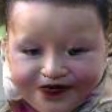}\includegraphics[width=0.06\linewidth]{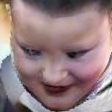}\includegraphics[width=0.06\linewidth]{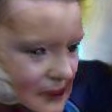}\includegraphics[width=0.06\linewidth]{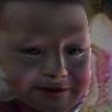}\includegraphics[width=0.06\linewidth]{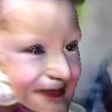} \\

\includegraphics[width=0.06\linewidth]{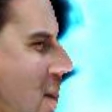}\includegraphics[width=0.06\linewidth]{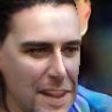}\includegraphics[width=0.06\linewidth]{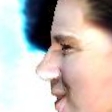}\includegraphics[width=0.06\linewidth]{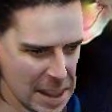}\includegraphics[width=0.06\linewidth]{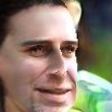}\includegraphics[width=0.06\linewidth]{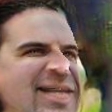}\includegraphics[width=0.06\linewidth]{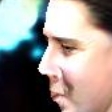}
&
\includegraphics[width=0.06\linewidth]{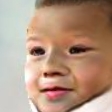}\includegraphics[width=0.06\linewidth]{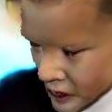}\includegraphics[width=0.06\linewidth]{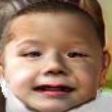}\includegraphics[width=0.06\linewidth]{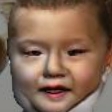}\includegraphics[width=0.06\linewidth]{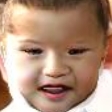}\includegraphics[width=0.06\linewidth]{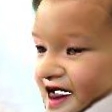}\includegraphics[width=0.06\linewidth]{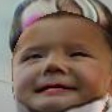}
\end{tabular}
   \vspace*{3mm}
   \caption{Random instances of a selection of IDs generated by the 3D GAN trained on FFHQ. Each row of seven images represents the same identity with random pose, expression and lighting. The images have been cropped to $112\times 112$ pixels for use in the experiments recorded in Table \ref{tab:CFPW_LFW_accuracies}.}
    \label{fig:FFHQ_random}
\end{figure*}

\begin{figure}
\begin{center}
\setlength\tabcolsep{4pt}
\begin{tabular}{cccc}
\includegraphics[width=0.21\linewidth]{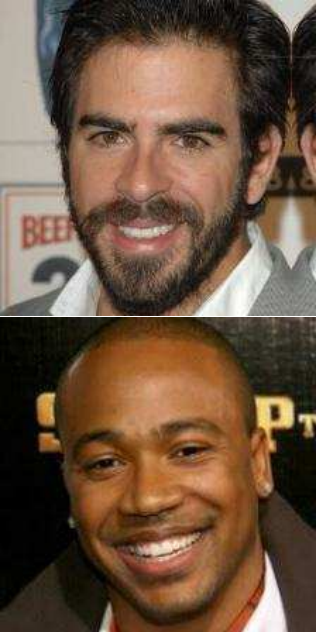} & \includegraphics[width=0.172\linewidth]{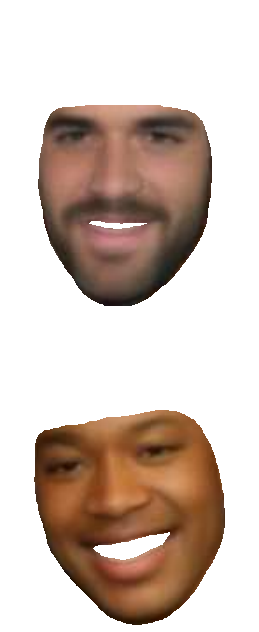} &
\includegraphics[width=0.172\linewidth]{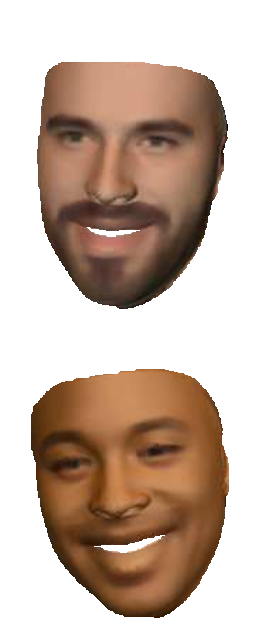} &
\includegraphics[width=0.172\linewidth]{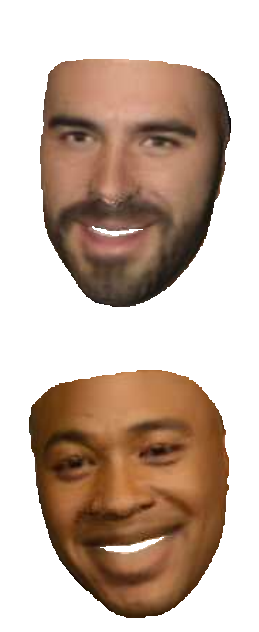} \\
Input & Tewari'18 \cite{DBLP:conf/cvpr/TewariZ0BKPT18} & Tran'18 \cite{DBLP:conf/cvpr/Tran018} & Tran'19 \cite{DBLP:conf/cvpr/TranL019}
\end{tabular}
\end{center}
   \caption{Example 3D facial reconstructions for three disentangled auto-encoder methods with texture models learnt from in-the-wild images. Images taken from \cite{DBLP:conf/cvpr/TranL019}.}
\label{fig:Qualitatve_comp}
\end{figure}

\subsubsection{Training datasets}
\label{sec:Training sets}

Our baseline FR experiments are trained on either CASIA Webface \cite{DBLP:journals/corr/YiLLL14a}, MS1M-V3 \cite{DBLP:journals/corr/abs-1905-00641} or our in-house dataset of 3.5 million images scraped from the internet, labelled as ``NetScrape'' in Figure \ref{fig:Pose_dists} and Tables \ref{tab:Dataset_comparison} and \ref{tab:Accuracies}. These datasets were then augmented using the 3D GAN trained on either CelebA \cite{DBLP:conf/iccv/LiuLWT15} or Flickr-Faces-HQ (FFHQ) \cite{DBLP:conf/cvpr/KarrasLA19}. Since CelebA is a dataset of potential benefit to FR, it was included in additional baseline experiments to provide a cleaner comparison where the dataset was used to train the 3D GAN. Details of these datasets are presented in Table \ref{tab:Dataset_comparison}. In Figure \ref{fig:Pose_dists} we show the relative distributions of yaw and pitch angles estimated using an in-house pose-estimation network. CelebA was found to have the narrowest ranges of both yaw and pitch. Despite this, in conjunction with the 3D GAN, we were able to use the dataset to improve large-pose facial recognition. CASIA Webface displays a noticeably wider distribution of yaw angles than the other datasets. Again, despite this prior advantage, we were able to improve FR results above the CASIA baselines.

Synthetic datasets of either 10k, 20k or 30k IDs were generated, each with 120 images per ID. Yaw and pitch angles were selected randomly from uniform distributions with ranges $[-90^{\circ}, 90^{\circ}]$ and $[-45^{\circ}, 45^{\circ}]$ respectively, whereas all other parameters (shape, expression, texture and background) were selected from a standard normal distribution, as during training. Synthetic images were augmented further using an SH lighting model \cite{DBLP:conf/siggraph/RamamoorthiH01a}. We augmented using only white light and chose ambient and non-ambient lighting coefficients from random uniform distributions in the ranges $[0.6, 1.4]$ and $[-0.4, 0.4]$ respectively. In performing this lighting augmentation, we make the assumption that images in the synthetic training dataset are only ambiently lit. This is not the case, however, and learned textures can contain problematic, embedded lighting effects. Nevertheless, performing this relatively crude lighting augmentation is shown to improve FR accuracy.

Examples of in-the-wild synthetic images can be seen in Figure \ref{fig:FFHQ_random}. These examples were generated from FFHQ at a resolution of $128\times 128$ pixels and then cropped to $112\times 112$ for use in the data-augmentation experiments recorded in Table \ref{tab:CFPW_LFW_accuracies}. The images are generally of lower quality than those generated from Multi-PIE and display visible artefacts. We suspect that this is due to a combination of the larger variation in textures and lighting conditions in FFHQ, the lower number of images at large poses, and the absence of reliable pose labels. Despite these issues, our experiments show that the synthetic data is of adequate quality to successfully augment FR datasets.

\begin{table*}[t]
\begin{center}
\resizebox{\textwidth}{!}{%
\begin{tabular}{|c|l|l|l|l|l|l|c|c|}
\hline
Exp & FR network & Loss & Training set & Augmentation & IDs & Images & CFP-FP & CPLFW \\
\hline\hline
1 & ResNet-34 & ArcFace & NetScrape & - & 26.8k & 3.5M & 93.59\% & 84.56\% \\
2 & ResNet-34 & ArcFace & NetScrape & CelebA & 26.8k + 10.2k & 3.5M + 0.2M & 94.06\% & 84.81\% \\
3 & ResNet-34 & ArcFace & NetScrape & 3D Synth (no SH) & 26.8k + 10k & 3.5M + 1.2M & 94.46\% & 85.55\% \\
4 & ResNet-34 & ArcFace & NetScrape & 3D Synth (narrow pose) & 26.8k + 10k & 3.5M + 1.2M & 93.76\% & 84.93\% \\
5 & ResNet-34 & ArcFace & NetScrape & 3D Synth & 26.8k + 10k & 3.5M + 1.2M & 94.89\% & \textbf{86.25\%} \\
6 & ResNet-34 & ArcFace & NetScrape & 3D Synth & 26.8k + 20k & 3.5M + 2.4M & \textbf{95.29\%} & 85.96\% \\
7 & ResNet-34 & ArcFace & NetScrape & 3D Synth & 26.8k + 30k & 3.5M + 3.6M & 94.63\% & 85.91\% \\
\hline
\end{tabular}
}
\end{center}
\caption{A comparison of the effect of augmentation with 3D GAN data (trained using CelebA) on CFP \cite{DBLP:conf/wacv/SenguptaCCPCJ16} and CPLFW \cite{zheng2018cross} verification accuracies . ``no SH'' indicates deactivation of SH lighting augmentation, and ``narrow pose'' indicates use of a Gaussian pose distribution of StdDev = $12^{\circ}$, similar to the distribution in CelebA.}
\label{tab:Accuracies}
\end{table*}

\begin{table*}[t]
\begin{center}
\resizebox{\textwidth}{!}{%
\begin{tabular}{|l|l|l|l|l|c|c|c|c|}
\hline
Method & FR network & Loss & Training set & Augmentation & CFP-FF & CFP-FP & LFW & CPLFW \\
\hline\hline
\rowcolor{Gainsboro}
Human & Brain & - & - & - & 96.24\% & 94.57\% & 97.27\% & 81.21\% \\
\hline\hline
\rowcolor{Gainsboro}
Baseline & ResNet-50 & ArcFace & CASIA & - & - & 95.56\% & - & - \\
\rowcolor{Gainsboro}
Gecer et al. (2019) \cite{DBLP:conf/eccv/GecerLPDPMZ20} & ResNet-50 & ArcFace & CASIA & 10k synth IDs & - & 97.12\% & - & - \\
\hline\hline
Baseline & ResNet-50 & ArcFace & CASIA & - & 99.37\% & 95.50\% & 99.30\% & 85.69\% \\
3D GAN (FFHQ) & ResNet-50 & ArcFace & CASIA & 10k synth IDs & 99.49\% & 96.40\% & 99.35\% & 86.53\% \\
3D GAN (FFHQ) & ResNet-50 & ArcFace & CASIA & 20k synth IDs & 99.40\% & 96.74\% & 99.42\% & 86.85\% \\
\hline\hline
\rowcolor{Gainsboro}
Deng et al. (2019) \cite{DBLP:conf/cvpr/DengGXZ19} & ResNet-100 & ArcFace & MS1M-V2 & - & - & - & 99.82\% & 92.08\% \\
\rowcolor{Gainsboro}
Huang et al. (2020a) \cite{DBLP:conf/eccv/HuangSTL0LHJ20} & ResNet-100 & ArcFace + DDL & VGG & - & - & \textbf{98.53\%} & 99.68\% & 93.43\% \\
\hline\hline
Baseline & ResNet-100 & ArcFace & MS1M-V3 & - & \textbf{99.90\%} & 98.47\% & \textbf{99.87\%} & 93.36\% \\
3D GAN (FFHQ) & ResNet-100 & ArcFace & MS1M-V3 & 20k synth IDs & \textbf{99.90\%} & 98.51\% & 99.85\% & \textbf{93.53\%} \\
\hline
\end{tabular}
}
\end{center}
\caption{A comparison of data-augmentation using synthetic identities generated by the 3D GAN (trained using FFHQ) with results from the literature (highlighted in grey). Evaluation is performed for the frontal-frontal (FF) and frontal-profile (FP) protocols of the CFP dataset as well as for LFW (view 2) and CPLFW. ``DLL'' refers to the Distribution Distillation Loss of \cite{DBLP:conf/eccv/HuangSTL0LHJ20}.}
\label{tab:CFPW_LFW_accuracies}
\end{table*}

In Figure \ref{fig:Qualitatve_comp} we provide example 3D facial reconstructions taken from \cite{DBLP:conf/cvpr/TranL019} for qualitative comparison with the images of Figure \ref{fig:FFHQ_random}. Despite the comparison being indirect, we believe 3D GAN images to be of a similar quality to the comparable state of the art. The 3D GAN also has the benefit of being able to 1) easily generate new identities, 2) generate full facial images, including the back of the head and the background, and 3) does not require the 3DMM to be fit to training images.

\subsubsection{Data-augmentation experiments}
\label{sec:Experiments}

In all of our experiments we use the ResNet architecture of \cite{DBLP:conf/cvpr/DengGXZ19} trained for 15 epochs. The only changes made were to the number of layers, as noted in Tables \ref{tab:Accuracies} and \ref{tab:CFPW_LFW_accuracies}. Table \ref{tab:Accuracies} presents results for a series of experiments in which we augmented the NetScrape dataset with 3D GAN data generated from CelebA. Experiment 1 gives our baseline, trained only on the ``NetScrape'' dataset. Experiment 2 shows that the effect of adding in CelebA is to increase accuracy on CFP-FP and CPLFW by 0.47\% and 0.25\% respectively. The effect of adding the synthetic data, however, is to increase accuracy by up to 1.7\% for CFP-FP, with an accuracy of 95.29\% being achieved in Experiment 6, and by up to 1.69\% for CPLFW, with an accuracy of 86.25\% being achieved in Experiment 5; i.e. the 3D GAN was able to exploit the images of CelebA somewhere between three to six times more effectively. Experiments 3 and 4 show that disabling the SH lighting, and limiting the variance of the pose to that found in CelebA itself, each decrease accuracy on both CFP-FP and CPLFW, with limitation of the pose having the largest effect. Both experiments, however, still perform better than the baseline. Finally, in Experiments 5, 6 and 7, we augment the dataset with 10k, 20k and 30k synthetic identities. For each experiment the measured accuracies are above those of the baseline experiments, although performance drops for either 20k or 30k identities depending on the evaluation dataset. The reason for this decrease in performance could be due to synthetic identities being too densely sampled, with too many lookalikes being generated. Alternatively, it could be due to overfitting of the biometric network to 3D GAN data since, in Experiments 6 and 7, significant proportions of the training dataset were synthetic ($40.7\%$ and $50.7\%$ as opposed to only $25.5\%$ in Experiment 3).

Table \ref{tab:CFPW_LFW_accuracies} presents the results of experiments for comparison with the 3D model-based data-augmentation method of \cite{DBLP:conf/eccv/GecerLPDPMZ20}, and with other results from the literature, highlighted in grey. The cleanest comparison is with the method of \cite{DBLP:conf/eccv/GecerLPDPMZ20} in which synthetic data generated by their TB-GAN was used to augment CASIA Webface giving an improvement of 1.56\% from 95.56\% to 97.12\% verification accuracy on CFP-FP. Augmentation using 20k synthetic identities generated from FFHQ using our 3D GAN gave an improvement of 1.24\% from the slightly lower baseline of accuracy of 95.50\% up to 96.74\%. Note that, in this experiment, the 3D GAN extracts useful information from the noisy FFHQ dataset, whereas the TB-GAN of \cite{DBLP:conf/eccv/GecerLPDPMZ20} is trained using a dataset of high-quality texture scans. Improvements in accuracy were also seen for CPLFW with addition of 10k and 20k synthetic identities leading to improvements of $0.84\%$ and $1.16\%$ respectively. Evaluation on the frontal protocol of CFP and on LFW gave only small improvements.

Finally, experiments were performed for a ResNet-100 architecture trained on the MS1M-V3 dataset. Augmentation using 20k synthetic identities generated from FFHQ using our 3D GAN gives a state-of-the-art accuracy of 93.53\% on CPLFW. This improvement was achieved despite the already very high performance of the baseline network.

\section{Conclusions}
\label{sec:Conclusions}

We proposed a novel 3D GAN formulation for learning a nonlinear texture model from in-the-wild images and thereby generating synthetic images of new identities with fully disentangled pose. We demonstrate that images synthesised by our 3D GAN can be used successfully to improve the accuracy of large-pose facial recognition. The 3D GAN enjoys the advantage of not requiring specially captured texture scans. Finally, since the 3D GAN can generate images of new identities, it provides an avenue for extraction of useful information from noisy datasets such as FFHQ.

{\small
\bibliographystyle{ieee_fullname}
\bibliography{egbib}

\begin{thebibliography}{10}\itemsep=-1pt

\bibitem{DBLP:conf/icml/ArjovskyCB17}
Mart{\'{\i}}n Arjovsky, Soumith Chintala, and L{\'{e}}on Bottou.
\newblock Wasserstein generative adversarial networks.
\newblock In Doina Precup and Yee~Whye Teh, editors, {\em Proceedings of the
  34th International Conference on Machine Learning, {ICML} 2017, Sydney, NSW,
  Australia, 6-11 August 2017}, volume~70 of {\em Proceedings of Machine
  Learning Research}, pages 214--223. {PMLR}, 2017.

\bibitem{DBLP:conf/siggraph/BlanzV99}
Volker Blanz and Thomas Vetter.
\newblock A morphable model for the synthesis of 3d faces.
\newblock In Warren~N. Waggenspack, editor, {\em Proceedings of the 26th Annual
  Conference on Computer Graphics and Interactive Techniques, {SIGGRAPH} 1999,
  Los Angeles, CA, USA, August 8-13, 1999}, pages 187--194. {ACM}, 1999.

\bibitem{DBLP:conf/cvpr/BoothAPTPZ17}
James Booth, Epameinondas Antonakos, Stylianos Ploumpis, George Trigeorgis,
  Yannis Panagakis, and Stefanos Zafeiriou.
\newblock 3d face morphable models "in-the-wild".
\newblock In {\em 2017 {IEEE} Conference on Computer Vision and Pattern
  Recognition, {CVPR} 2017, Honolulu, HI, USA, July 21-26, 2017}, pages
  5464--5473. {IEEE} Computer Society, 2017.

\bibitem{DBLP:conf/cvpr/BoothRZPD16}
James Booth, Anastasios Roussos, Stefanos Zafeiriou, Allan Ponniah, and David
  Dunaway.
\newblock A 3d morphable model learnt from 10, 000 faces.
\newblock In {\em 2016 {IEEE} Conference on Computer Vision and Pattern
  Recognition, {CVPR} 2016, Las Vegas, NV, USA, June 27-30, 2016}, pages
  5543--5552. {IEEE} Computer Society, 2016.

\bibitem{DBLP:conf/cvpr/CaoR0TL18}
Kaidi Cao, Yu Rong, Cheng Li, Xiaoou Tang, and Chen~Change Loy.
\newblock Pose-robust face recognition via deep residual equivariant mapping.
\newblock In {\em 2018 {IEEE} Conference on Computer Vision and Pattern
  Recognition, {CVPR} 2018, Salt Lake City, UT, USA, June 18-22, 2018}, pages
  5187--5196. {IEEE} Computer Society, 2018.

\bibitem{DBLP:conf/eccv/ChaudhuriVSW20}
Bindita Chaudhuri, Noranart Vesdapunt, Linda~G. Shapiro, and Baoyuan Wang.
\newblock Personalized face modeling for improved face reconstruction and
  motion retargeting.
\newblock In Andrea Vedaldi, Horst Bischof, Thomas Brox, and Jan{-}Michael
  Frahm, editors, {\em Computer Vision - {ECCV} 2020 - 16th European
  Conference, Glasgow, UK, August 23-28, 2020, Proceedings, Part {V}}, volume
  12350 of {\em Lecture Notes in Computer Science}, pages 142--160. Springer,
  2020.

\bibitem{DBLP:journals/corr/CrispellBCBM17}
Daniel~E. Crispell, Octavian Biris, Nate Crosswhite, Jeffrey Byrne, and
  Joseph~L. Mundy.
\newblock Dataset augmentation for pose and lighting invariant face
  recognition.
\newblock {\em CoRR}, abs/1704.04326, 2017.

\bibitem{DBLP:conf/cvpr/DengCXZZ18}
Jiankang Deng, Shiyang Cheng, Niannan Xue, Yuxiang Zhou, and Stefanos
  Zafeiriou.
\newblock {UV-GAN:} adversarial facial {UV} map completion for pose-invariant
  face recognition.
\newblock In {\em 2018 {IEEE} Conference on Computer Vision and Pattern
  Recognition, {CVPR} 2018, Salt Lake City, UT, USA, June 18-22, 2018}, pages
  7093--7102. {IEEE} Computer Society, 2018.

\bibitem{DBLP:conf/cvpr/DengGXZ19}
Jiankang Deng, Jia Guo, Niannan Xue, and Stefanos Zafeiriou.
\newblock Arcface: Additive angular margin loss for deep face recognition.
\newblock In {\em {IEEE} Conference on Computer Vision and Pattern Recognition,
  {CVPR} 2019, Long Beach, CA, USA, June 16-20, 2019}, pages 4690--4699.
  Computer Vision Foundation / {IEEE}, 2019.

\bibitem{DBLP:journals/corr/abs-1905-00641}
Jiankang Deng, Jia Guo, Yuxiang Zhou, Jinke Yu, Irene Kotsia, and Stefanos
  Zafeiriou.
\newblock Retinaface: Single-stage dense face localisation in the wild.
\newblock {\em CoRR}, abs/1905.00641, 2019.

\bibitem{DBLP:conf/eccv/GecerBKK18}
Baris Gecer, Binod Bhattarai, Josef Kittler, and Tae{-}Kyun Kim.
\newblock Semi-supervised adversarial learning to generate photorealistic face
  images of new identities from 3d morphable model.
\newblock In Vittorio Ferrari, Martial Hebert, Cristian Sminchisescu, and Yair
  Weiss, editors, {\em Computer Vision - {ECCV} 2018 - 15th European
  Conference, Munich, Germany, September 8-14, 2018, Proceedings, Part {XI}},
  volume 11215 of {\em Lecture Notes in Computer Science}, pages 230--248.
  Springer, 2018.

\bibitem{DBLP:conf/eccv/GecerLPDPMZ20}
Baris Gecer, Alexandros Lattas, Stylianos Ploumpis, Jiankang Deng, Athanasios
  Papaioannou, Stylianos Moschoglou, and Stefanos Zafeiriou.
\newblock Synthesizing coupled 3d face modalities by trunk-branch generative
  adversarial networks.
\newblock In Andrea Vedaldi, Horst Bischof, Thomas Brox, and Jan{-}Michael
  Frahm, editors, {\em Computer Vision - {ECCV} 2020 - 16th European
  Conference, Glasgow, UK, August 23-28, 2020, Proceedings, Part {XXIX}},
  volume 12374 of {\em Lecture Notes in Computer Science}, pages 415--433.
  Springer, 2020.

\bibitem{DBLP:conf/cvpr/GecerPKZ19}
Baris Gecer, Stylianos Ploumpis, Irene Kotsia, and Stefanos Zafeiriou.
\newblock {GANFIT:} generative adversarial network fitting for high fidelity 3d
  face reconstruction.
\newblock In {\em {IEEE} Conference on Computer Vision and Pattern Recognition,
  {CVPR} 2019, Long Beach, CA, USA, June 16-20, 2019}, pages 1155--1164.
  Computer Vision Foundation / {IEEE}, 2019.

\bibitem{DBLP:conf/3dim/GhoshGURBB20}
Partha Ghosh, Pravir~Singh Gupta, Roy Uziel, Anurag Ranjan, Michael~J. Black,
  and Timo Bolkart.
\newblock {GIF:} generative interpretable faces.
\newblock In Vitomir Struc and Francisco~G{\'{o}}mez Fern{\'{a}}ndez, editors,
  {\em 8th International Conference on 3D Vision, 3DV 2020, Virtual Event,
  Japan, November 25-28, 2020}, pages 868--878. {IEEE}, 2020.

\bibitem{DBLP:journals/ivc/GrossMCKB10}
Ralph Gross, Iain~A. Matthews, Jeffrey~F. Cohn, Takeo Kanade, and Simon Baker.
\newblock Multi-pie.
\newblock {\em Image Vis. Comput.}, 28(5):807--813, 2010.

\bibitem{DBLP:conf/nips/GulrajaniAADC17}
Ishaan Gulrajani, Faruk Ahmed, Mart{\'{\i}}n Arjovsky, Vincent Dumoulin, and
  Aaron~C. Courville.
\newblock Improved training of wasserstein gans.
\newblock In Isabelle Guyon, Ulrike von Luxburg, Samy Bengio, Hanna~M. Wallach,
  Rob Fergus, S.~V.~N. Vishwanathan, and Roman Garnett, editors, {\em Advances
  in Neural Information Processing Systems 30: Annual Conference on Neural
  Information Processing Systems 2017, 4-9 December 2017, Long Beach, CA,
  {USA}}, pages 5767--5777, 2017.

\bibitem{DBLP:conf/cvpr/HassnerHPE15}
Tal Hassner, Shai Harel, Eran Paz, and Roee Enbar.
\newblock Effective face frontalization in unconstrained images.
\newblock In {\em {IEEE} Conference on Computer Vision and Pattern Recognition,
  {CVPR} 2015, Boston, MA, USA, June 7-12, 2015}, pages 4295--4304. {IEEE}
  Computer Society, 2015.

\bibitem{DBLP:journals/ijcv/HendersonF20}
Paul Henderson and Vittorio Ferrari.
\newblock Learning single-image 3d reconstruction by generative modelling of
  shape, pose and shading.
\newblock {\em Int. J. Comput. Vis.}, 128(4):835--854, 2020.

\bibitem{DBLP:conf/eccv/HuangSTL0LHJ20}
Yuge Huang, Pengcheng Shen, Ying Tai, Shaoxin Li, Xiaoming Liu, Jilin Li,
  Feiyue Huang, and Rongrong Ji.
\newblock Improving face recognition from hard samples via distribution
  distillation loss.
\newblock In Andrea Vedaldi, Horst Bischof, Thomas Brox, and Jan{-}Michael
  Frahm, editors, {\em Computer Vision - {ECCV} 2020 - 16th European
  Conference, Glasgow, UK, August 23-28, 2020, Proceedings, Part {XXX}}, volume
  12375 of {\em Lecture Notes in Computer Science}, pages 138--154. Springer,
  2020.

\bibitem{DBLP:conf/cvpr/HuangWT0SLLH20}
Yuge Huang, Yuhan Wang, Ying Tai, Xiaoming Liu, Pengcheng Shen, Shaoxin Li,
  Jilin Li, and Feiyue Huang.
\newblock Curricularface: Adaptive curriculum learning loss for deep face
  recognition.
\newblock In {\em 2020 {IEEE/CVF} Conference on Computer Vision and Pattern
  Recognition, {CVPR} 2020, Seattle, WA, USA, June 13-19, 2020}, pages
  5900--5909. {IEEE}, 2020.

\bibitem{DBLP:conf/iclr/KarrasALL18}
Tero Karras, Timo Aila, Samuli Laine, and Jaakko Lehtinen.
\newblock Progressive growing of gans for improved quality, stability, and
  variation.
\newblock In {\em 6th International Conference on Learning Representations,
  {ICLR} 2018, Vancouver, BC, Canada, April 30 - May 3, 2018, Conference Track
  Proceedings}. OpenReview.net, 2018.

\bibitem{DBLP:conf/cvpr/KarrasLA19}
Tero Karras, Samuli Laine, and Timo Aila.
\newblock A style-based generator architecture for generative adversarial
  networks.
\newblock In {\em {IEEE} Conference on Computer Vision and Pattern Recognition,
  {CVPR} 2019, Long Beach, CA, USA, June 16-20, 2019}, pages 4401--4410.
  Computer Vision Foundation / {IEEE}, 2019.

\bibitem{DBLP:conf/cvpr/KarrasLAHLA20}
Tero Karras, Samuli Laine, Miika Aittala, Janne Hellsten, Jaakko Lehtinen, and
  Timo Aila.
\newblock Analyzing and improving the image quality of stylegan.
\newblock In {\em 2020 {IEEE/CVF} Conference on Computer Vision and Pattern
  Recognition, {CVPR} 2020, Seattle, WA, USA, June 13-19, 2020}, pages
  8107--8116. {IEEE}, 2020.

\bibitem{DBLP:journals/corr/abs-1802-05891}
Adam Kortylewski, Andreas Schneider, Thomas Gerig, Bernhard Egger, Andreas
  Morel{-}Forster, and Thomas Vetter.
\newblock Training deep face recognition systems with synthetic data.
\newblock {\em CoRR}, abs/1802.05891, 2018.

\bibitem{DBLP:conf/eccv/KowalskiGEBJS20}
Marek Kowalski, Stephan~J. Garbin, Virginia Estellers, Tadas Baltrusaitis,
  Matthew Johnson, and Jamie Shotton.
\newblock {CONFIG:} controllable neural face image generation.
\newblock In Andrea Vedaldi, Horst Bischof, Thomas Brox, and Jan{-}Michael
  Frahm, editors, {\em Computer Vision - {ECCV} 2020 - 16th European
  Conference, Glasgow, UK, August 23-28, 2020, Proceedings, Part {XI}}, volume
  12356 of {\em Lecture Notes in Computer Science}, pages 299--315. Springer,
  2020.

\bibitem{DBLP:journals/tog/LiBBL017}
Tianye Li, Timo Bolkart, Michael~J. Black, Hao Li, and Javier Romero.
\newblock Learning a model of facial shape and expression from 4d scans.
\newblock {\em {ACM} Trans. Graph.}, 36(6):194:1--194:17, 2017.

\bibitem{DBLP:conf/iccv/LiuLWT15}
Ziwei Liu, Ping Luo, Xiaogang Wang, and Xiaoou Tang.
\newblock Deep learning face attributes in the wild.
\newblock In {\em 2015 {IEEE} International Conference on Computer Vision,
  {ICCV} 2015, Santiago, Chile, December 7-13, 2015}, pages 3730--3738. {IEEE}
  Computer Society, 2015.

\bibitem{DBLP:journals/ijon/LvSHZZ17}
Jiang{-}Jing Lv, Xiaohu Shao, Jia{-}Shui Huang, Xiang{-}Dong Zhou, and Xi Zhou.
\newblock Data augmentation for face recognition.
\newblock {\em Neurocomputing}, 230:184--196, 2017.

\bibitem{marriott2020assessment}
Richard Marriott, Safa Madiouni, Sami Romdhani, Stephane Gentric, and Liming
  Chen.
\newblock An assessment of {GAN}s for identity-related applications.
\newblock In {\em 2020 IEEE International Joint Conference on Biometrics
  ({IJCB})}. IEEE, 2020.

\bibitem{DBLP:conf/eccv/MasiTHLM16}
Iacopo Masi, Anh~Tuan Tran, Tal Hassner, Jatuporn~Toy Leksut, and
  G{\'{e}}rard~G. Medioni.
\newblock Do we really need to collect millions of faces for effective face
  recognition?
\newblock In Bastian Leibe, Jiri Matas, Nicu Sebe, and Max Welling, editors,
  {\em Computer Vision - {ECCV} 2016 - 14th European Conference, Amsterdam, The
  Netherlands, October 11-14, 2016, Proceedings, Part {V}}, volume 9909 of {\em
  Lecture Notes in Computer Science}, pages 579--596. Springer, 2016.

\bibitem{DBLP:conf/iccv/PengYSMC17}
Xi Peng, Xiang Yu, Kihyuk Sohn, Dimitris~N. Metaxas, and Manmohan Chandraker.
\newblock Reconstruction-based disentanglement for pose-invariant face
  recognition.
\newblock In {\em {IEEE} International Conference on Computer Vision, {ICCV}
  2017, Venice, Italy, October 22-29, 2017}, pages 1632--1641. {IEEE} Computer
  Society, 2017.

\bibitem{DBLP:conf/siggraph/RamamoorthiH01a}
Ravi Ramamoorthi and Pat Hanrahan.
\newblock An efficient representation for irradiance environment maps.
\newblock In Lynn Pocock, editor, {\em Proceedings of the 28th Annual
  Conference on Computer Graphics and Interactive Techniques, {SIGGRAPH} 2001,
  Los Angeles, California, USA, August 12-17, 2001}, pages 497--500. {ACM},
  2001.

\bibitem{DBLP:conf/wacv/SenguptaCCPCJ16}
Soumyadip Sengupta, Jun{-}Cheng Chen, Carlos~Domingo Castillo, Vishal~M. Patel,
  Rama Chellappa, and David~W. Jacobs.
\newblock Frontal to profile face verification in the wild.
\newblock In {\em 2016 {IEEE} Winter Conference on Applications of Computer
  Vision, {WACV} 2016, Lake Placid, NY, USA, March 7-10, 2016}, pages 1--9.
  {IEEE} Computer Society, 2016.

\bibitem{DBLP:conf/cvpr/Shen0YWT18}
Yujun Shen, Ping Luo, Junjie Yan, Xiaogang Wang, and Xiaoou Tang.
\newblock Faceid-gan: Learning a symmetry three-player {GAN} for
  identity-preserving face synthesis.
\newblock In {\em 2018 {IEEE} Conference on Computer Vision and Pattern
  Recognition, {CVPR} 2018, Salt Lake City, UT, USA, June 18-22, 2018}, pages
  821--830. {IEEE} Computer Society, 2018.

\bibitem{DBLP:conf/cvpr/TewariB0BESPZT19}
Ayush Tewari, Florian Bernard, Pablo Garrido, Gaurav Bharaj, Mohamed Elgharib,
  Hans{-}Peter Seidel, Patrick P{\'{e}}rez, Michael Zollh{\"{o}}fer, and
  Christian Theobalt.
\newblock {FML:} face model learning from videos.
\newblock In {\em {IEEE} Conference on Computer Vision and Pattern Recognition,
  {CVPR} 2019, Long Beach, CA, USA, June 16-20, 2019}, pages 10812--10822.
  Computer Vision Foundation / {IEEE}, 2019.

\bibitem{DBLP:conf/cvpr/TewariEBBSPZT20}
Ayush Tewari, Mohamed Elgharib, Gaurav Bharaj, Florian Bernard, Hans{-}Peter
  Seidel, Patrick P{\'{e}}rez, Michael Zollh{\"{o}}fer, and Christian Theobalt.
\newblock Stylerig: Rigging stylegan for 3d control over portrait images.
\newblock In {\em 2020 {IEEE/CVF} Conference on Computer Vision and Pattern
  Recognition, {CVPR} 2020, Seattle, WA, USA, June 13-19, 2020}, pages
  6141--6150. {IEEE}, 2020.

\bibitem{DBLP:journals/tog/TewariE0BSPZT20}
Ayush Tewari, Mohamed Elgharib, Mallikarjun~B. R., Florian Bernard,
  Hans{-}Peter Seidel, Patrick P{\'{e}}rez, Michael Zollh{\"{o}}fer, and
  Christian Theobalt.
\newblock {PIE:} portrait image embedding for semantic control.
\newblock {\em {ACM} Trans. Graph.}, 39(6):223:1--223:14, 2020.

\bibitem{DBLP:conf/cvpr/TewariZ0BKPT18}
Ayush Tewari, Michael Zollh{\"{o}}fer, Pablo Garrido, Florian Bernard,
  Hyeongwoo Kim, Patrick P{\'{e}}rez, and Christian Theobalt.
\newblock Self-supervised multi-level face model learning for monocular
  reconstruction at over 250 hz.
\newblock In {\em 2018 {IEEE} Conference on Computer Vision and Pattern
  Recognition, {CVPR} 2018, Salt Lake City, UT, USA, June 18-22, 2018}, pages
  2549--2559. {IEEE} Computer Society, 2018.

\bibitem{DBLP:conf/cvpr/TranL019}
Luan Tran, Feng Liu, and Xiaoming Liu.
\newblock Towards high-fidelity nonlinear 3d face morphable model.
\newblock In {\em {IEEE} Conference on Computer Vision and Pattern Recognition,
  {CVPR} 2019, Long Beach, CA, USA, June 16-20, 2019}, pages 1126--1135.
  Computer Vision Foundation / {IEEE}, 2019.

\bibitem{DBLP:conf/cvpr/Tran018}
Luan Tran and Xiaoming Liu.
\newblock Nonlinear 3d face morphable model.
\newblock In {\em 2018 {IEEE} Conference on Computer Vision and Pattern
  Recognition, {CVPR} 2018, Salt Lake City, UT, USA, June 18-22, 2018}, pages
  7346--7355. {IEEE} Computer Society, 2018.

\bibitem{DBLP:conf/cvpr/Tran0017}
Luan Tran, Xi Yin, and Xiaoming Liu.
\newblock Disentangled representation learning {GAN} for pose-invariant face
  recognition.
\newblock In {\em 2017 {IEEE} Conference on Computer Vision and Pattern
  Recognition, {CVPR} 2017, Honolulu, HI, USA, July 21-26, 2017}, pages
  1283--1292. {IEEE} Computer Society, 2017.

\bibitem{DBLP:journals/corr/YiLLL14a}
Dong Yi, Zhen Lei, Shengcai Liao, and Stan~Z. Li.
\newblock Learning face representation from scratch.
\newblock {\em CoRR}, abs/1411.7923, 2014.

\bibitem{DBLP:conf/nips/ZhaoXKLZWPSYF17}
Jian Zhao, Lin Xiong, Jayashree Karlekar, Jianshu Li, Fang Zhao, Zhecan Wang,
  Sugiri Pranata, Shengmei Shen, Shuicheng Yan, and Jiashi Feng.
\newblock Dual-agent gans for photorealistic and identity preserving profile
  face synthesis.
\newblock In Isabelle Guyon, Ulrike von Luxburg, Samy Bengio, Hanna~M. Wallach,
  Rob Fergus, S.~V.~N. Vishwanathan, and Roman Garnett, editors, {\em Advances
  in Neural Information Processing Systems 30: Annual Conference on Neural
  Information Processing Systems 2017, 4-9 December 2017, Long Beach, CA,
  {USA}}, pages 66--76, 2017.

\bibitem{zheng2018cross}
Tianyue Zheng and Weihong Deng.
\newblock Cross-pose {LFW}: A database for studying cross-pose face recognition
  in unconstrained environments.
\newblock {\em Beijing University of Posts and Telecommunications, Tech. Rep},
  5, 2018.

\bibitem{DBLP:conf/cvpr/ZhuLYYL15}
Xiangyu Zhu, Zhen Lei, Junjie Yan, Dong Yi, and Stan~Z. Li.
\newblock High-fidelity pose and expression normalization for face recognition
  in the wild.
\newblock In {\em {IEEE} Conference on Computer Vision and Pattern Recognition,
  {CVPR} 2015, Boston, MA, USA, June 7-12, 2015}, pages 787--796. {IEEE}
  Computer Society, 2015.

\end{thebibliography}
}

\end{document}